\documentclass[sigconf]{acmart}

\AtBeginDocument{%
  \providecommand\BibTeX{{%
    \normalfont B\kern-0.5em{\scshape i\kern-0.25em b}\kern-0.8em\TeX}}}

\copyrightyear{2020}
\acmYear{2020}
\setcopyright{acmcopyright}\acmConference[MM '20]{Proceedings of the 28th ACM International Conference on Multimedia}{October 12--16, 2020}{Seattle, WA, USA}
\acmBooktitle{Proceedings of the 28th ACM International Conference on Multimedia (MM '20), October 12--16, 2020, Seattle, WA, USA}
\acmPrice{15.00}
\acmDOI{10.1145/3394171.3413518}
\acmISBN{978-1-4503-7988-5/20/10}
\usepackage{multirow}
\usepackage[utf8]{inputenc}

\begin{document}

%%
%% The "title" command has an optional parameter,
%% allowing the author to define a "short title" to be used in page headers.
\title{DeVLBert: Learning Deconfounded Visio-Linguistic Representations}

%
% The "author" command and its associated commands are used to define
% the authors and their affiliations.
% Of note is the shared affiliation of the first two authors, and the
% "authornote" and "authornotemark" commands
% used to denote shared contribution to the research.
\author[S. Zhang*, T. Jiang*, T. Wang, K. Kuang, Z. Zhao, J. Zhu, J. Yu, H. Yang, F. Wu]{
    Shengyu Zhang$^{1*}$, Tan Jiang$^{1*}$, Tan Wang$^{2}$, Kun Kuang$^{1\dagger}$, Zhou Zhao$^{1}$, Jianke Zhu$^{1}$, Jin Yu$^{3}$, Hongxia Yang$^{3\dagger}$, Fei Wu$^{1\dagger}$
}
\affiliation{
    $^1$ College of Computer Science and Technology, Zhejiang University
}
\affiliation{
	$^2$ University of Electronic Science and Technology of China
}
\affiliation{
	$^3$ Alibaba Group
}
\email{
  {sy_zhang, jiangtan, zhaozhou, kunkuang, wufei}@zju.edu.cn
}
\email{
  {kola.yu, yang.yhx}@alibaba-inc.com
}
\email{
	wangt97@hotmail.com
}
\renewcommand{\authors}{Shengyu Zhang, Tan Jiang, Tan Wang, Kun Kuang, Zhou Zhao, Jianke Zhu, Jin Yu, Hongxia Yang, Fei Wu}

%%
%% By default, the full list of authors will be used in the page
%% headers. Often, this list is too long, and will overlap
%% other information printed in the page headers. This command allows
%% the author to define a more concise list
%% of authors' names for this purpose.
%\renewcommand{\shortauthors}{S. Zhang*, Z. Tan*, J. Yu, Z. Zhao, K. Kuang, T. Jiang, J. Zhou, H. Yang, F. Wu}
\newcommand{\model}{ model }
\newcommand{\etal}{\textit{et al}.}
\newcommand{\ie}{\textit{i}.\textit{e}.}
\newcommand{\eg}{\textit{e}.\textit{g}.}
\newcommand{\vpara}[1]{\vspace{0.05in}\noindent\textbf{#1 }}

%%
%% The abstract is a short summary of the work to be presented in the
%% article.
\begin{abstract}

In this paper, we propose to investigate the problem of out-of-domain visio-linguistic pretraining, where the pretraining data distribution differs from that of downstream data on which the pretrained model will be fine-tuned. Existing methods for this problem are purely likelihood-based, leading to the spurious correlations and hurt the generalization ability when transferred to out-of-domain downstream tasks. By spurious correlation, we mean that the conditional probability of one token (object or word) given another one can be high (due to the dataset biases) without robust (causal) relationships between them. To mitigate such dataset biases, we propose a Deconfounded Visio-Linguistic Bert framework, abbreviated as DeVLBert, to perform intervention-based learning. We borrow the idea of the backdoor adjustment from the research field of causality and propose several neural-network based architectures for Bert-style out-of-domain pretraining. The quantitative results on three downstream tasks, Image Retrieval (IR), Zero-shot IR, and Visual Question Answering, show the effectiveness of DeVLBert by boosting generalization ability.

\end{abstract}

%%
%% The code below is generated by the tool at http://dl.acm.org/ccs.cfm.
%% Please copy and paste the code instead of the example below.
%%



%\begin{CCSXML}
%<ccs2012>
% <concept>
%  <concept_id>10010520.10010553.10010562</concept_id>
%  <concept_desc>Computer systems organization~Embedded systems</concept_desc>
%  <concept_significance>500</concept_significance>
% </concept>
% <concept>
%  <concept_id>10010520.10010575.10010755</concept_id>
%  <concept_desc>Computer systems organization~Redundancy</concept_desc>
%  <concept_significance>300</concept_significance>
% </concept>
% <concept>
%  <concept_id>10010520.10010553.10010554</concept_id>
%  <concept_desc>Computer systems organization~Robotics</concept_desc>
%  <concept_significance>100</concept_significance>
% </concept>
% <concept>
%  <concept_id>10003033.10003083.10003095</concept_id>
%  <concept_desc>Networks~Network reliability</concept_desc>
%  <concept_significance>100</concept_significance>
% </concept>
%</ccs2012>
%\end{CCSXML}
%
%\ccsdesc[500]{Computer systems organization~Embedded systems}
%\ccsdesc[300]{Computer systems organization~Redundancy}
%\ccsdesc{Computer systems organization~Robotics}
%\ccsdesc[100]{Networks~Network reliability}

%%
%% Keywords. The author(s) should pick words that accurately describe
%% the work being presented. Separate the keywords with commas.


%% A "teaser" image appears between the author and affiliation
%% information and the body of the document, and typically spans the
%% page.
%\begin{teaserfigure}
%  \includegraphics[width=\textwidth]{sampleteaser}
%  \caption{Seattle Mariners at Spring Training, 2010.}
%  \Description{Enjoying the baseball game from the third-base
%  seats. Ichiro Suzuki preparing to bat.}
%  \label{fig:teaser}
%\end{teaserfigure}

%%
%% This command processes the author and affiliation and title
%% information and builds the first part of the formatted document.
\maketitle

\renewcommand{\thefootnote}{\fnsymbol{footnote}}
\footnotetext[1]{These authors contributed equally to this work.}
\footnotetext[2]{Corresponding Authors.}
\footnotetext{Work was performed when S. Zhang and T. Jiang were interns at Alibaba Group.}
\renewcommand{\thefootnote}{\arabic{footnote}}
\fancyhead{}

\section{Introduction}

Since early attempts that pretrain a backbone model \cite{Krizhevsky_Sutskever_Hinton_2012, Simonyan_Zisserman_2015, He_Zhang_Ren_Sun_2016} on large-scale dataset \cite{Deng_Dong_Socher_Li_Li_Li_2009} and then transfer the knowledge to numerous computer vision tasks, pretraining has become a hallmark of the success of deep learning. More recently, the volume of transformer-based and Bert-style pretraining models \cite{Devlin_Chang_Lee_Toutanova_2019,Song_Tan_Qin_Lu_Liu_2019,Dong_Yang_Wang_Wei_Liu_Wang_Gao_Zhou_Hon_2019,Conneau_Lample_2019,Liu_Ott_Goyal_Du_Joshi_Chen_Levy_Lewis_Zettlemoyer_Stoyanov_2019} has grown tremendously in the research field of natural language processing and has achieved state-of-the-art performance in various NLP tasks. Likewise, the success of Bert-style pretraining techniques has been transferred to the research field of the intersection of vision and language \cite{Lu_Batra_Parikh_Lee_2019,Li_Yatskar_Yin_Hsieh_Chang_2019,Su_Zhu_Cao_Li_Lu_Wei_Dai_2020,Sun_Myers_Vondrick_Murphy_Schmid_2019,Sun_Baradel_Murphy_Schmid_2019}.

%--------------------------------fig-------------------------
\begin{figure}[!t] \begin{center}
    \includegraphics[width=0.9\columnwidth]{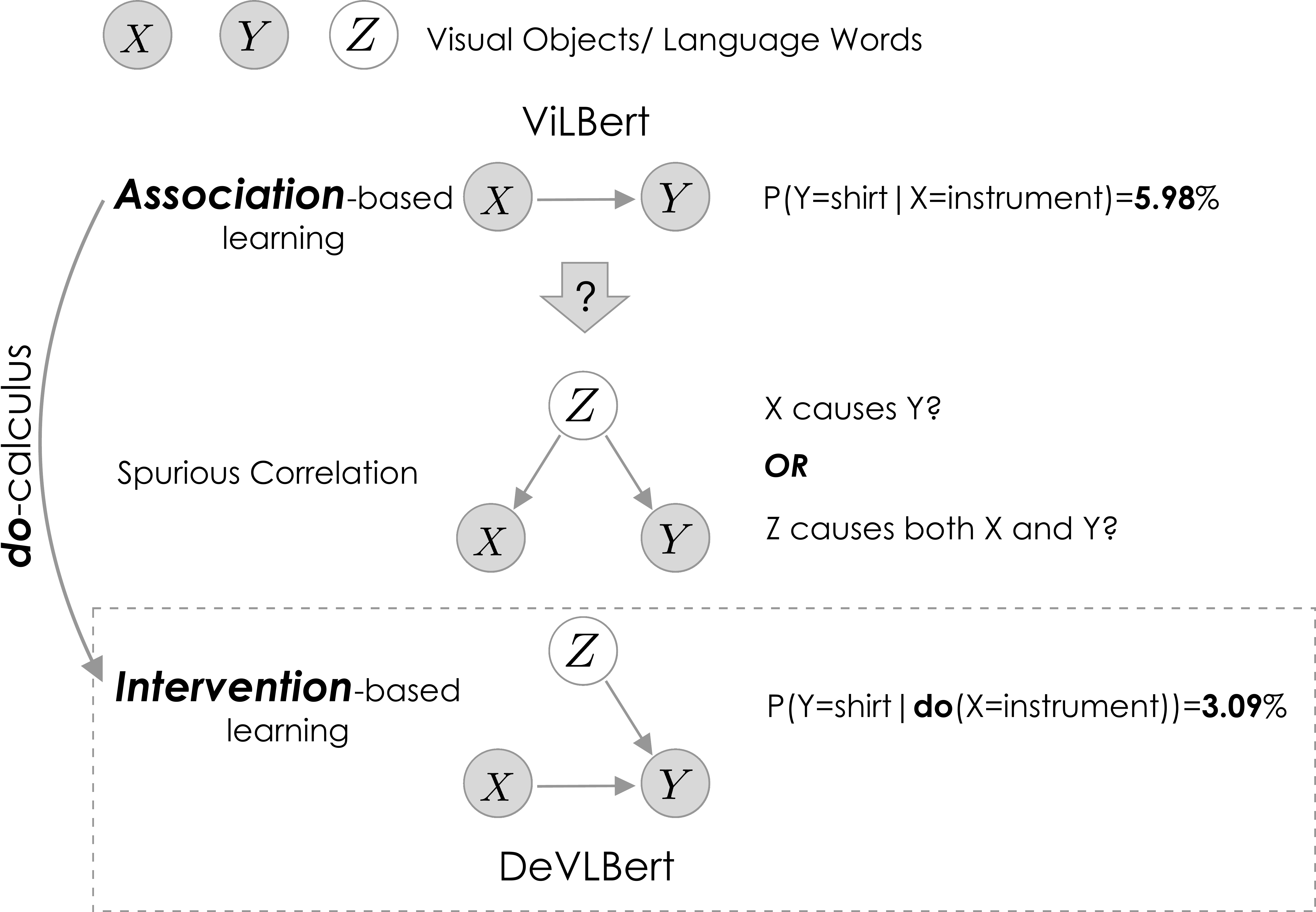}
    \caption{
    An illustration of the transition from traditional association-based learning to causal intervention-based learning. The critical difference is that the intervention mitigates the spurious correlation by blocking the back-door path $\protect Z \rightarrow X$ and thus controlling the condition $\protect X$.
    	}
\label{fig:first}
\end{center} 
\vspace{-0.4cm}
\end{figure}

%--------------------------------fig end--------------------

%--------------------------------fig---------------------
\begin{figure*}[t] \begin{center}
    \includegraphics[width=0.85\textwidth]{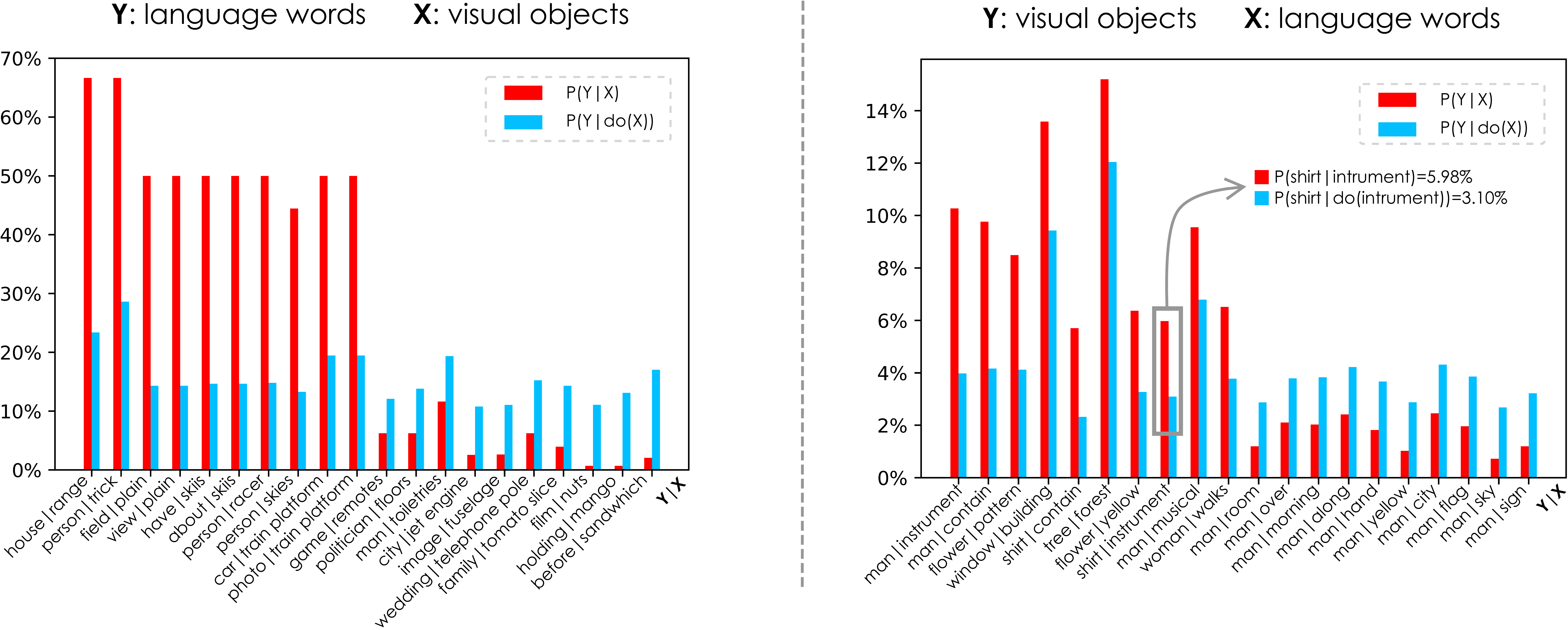}
    \caption{
    The conditional probabilities of a token Y (visual object or language word) given another token X (from the other modality) before intervention (\ie, P(Y|X), colored in red) and after intervention (\ie, P(Y|do(X), colored in blue). %We show the case that Y belongs to language words and Y belongs to language objects, and the opposite.
	}
\vspace{-0.2cm}
\label{fig:condi}
\end{center} \end{figure*}
%--------------------------------fig end---------------------

Despite the significant progress that recent methods have made over the initiative work ViLBert \cite{Lu_Batra_Parikh_Lee_2019}, part of their success can be traced back to the introduction of \textit{in-domain} pretraining datasets besides the Conceptual Caption \cite{Sharma_Ding_Goodman_Soricut_2018} dataset. By \textit{in-domain}, we refer to those datasets used in both pretraining and downstream tasks, such as MSCOCO \cite{Lin_Maire_Belongie_Hays_Perona_Ramanan_Dollr_Zitnick_2014}, and Visual Genome \cite{Krishna_Zhu_Groth_Johnson_Hata_Kravitz_Chen_Kalantidis_Li_Shamma_et_2017}. 
However, out-of-domain pretraining, \ie, pretraining models on \textit{out-of-domain} datasets and transferring the learned knowledge into downstream tasks with \textbf{unkown} data distributions, can be an essential research topic. 
In this paper, we focus on out-of-domain pretraining and learning generic representations as the ViLBert does.

A fundamental requirement for out-of-domain transfer learning is to mitigate the biases from the pretraining data \cite{Wang_Huang_Zhang_Sun_2020}, which may be useful for the in-domain testing but harmful for out-of-domain testing \cite{Kuang_Cui_Athey_Xiong_Li_2018} due to the \textit{spurious correlation} \cite{Pearl_Glymour_Jewell_2016}. 
To verify such existence of the correlation biases, we follow \cite{Wang_Huang_Zhang_Sun_2020} to conduct a toy experiment on Conceptual Caption dataset.
We observe that the conditional probability of \texttt{shirt} (visual object) given the \texttt{instrument} (language word) is large, \ie, $p(\textrm{shirt}|\textrm{instrument}) = 5.98\%$, but there are actually no robust relationships between them.
Most previous works just blame this for the biased data collection without further justification. 
However, this is not reasonable since we human ourselves are just living in a biased nature.
In our methodology, we draw inspiration from the causal inference \cite{kuang2020causal} and borrow the idea of the backdoor adjustment (also known as covariate adjustment or statistical adjustment) \cite{Wang_Huang_Zhang_Sun_2020, neuberg2003causality,angrist2008mostly} to mitigate these biases. 
As shown in Figure \ref{fig:first}, 
the traditional \textit{association-based} learning fashion may lead to the spurious correlation between two tokens (visual objects or language words) by a common cause, \ie, the confounder.
By introducing backdoor adjustment (deconfounding), the original conditional probability of $p(\textrm{shirt}|\textrm{instrument})$ can be adjusted to $3.10\%$ (nearly half) with a \textbf{do} operator. 
The essence of deconfounding is to control the condition (\texttt{instrument}) from being affected by other potential confounders when assessing the effect on the outcome (\texttt{shirt}) given the condition, \ie, \textbf{intervention}. 
In this way, the pure \textit{association}-based pretraining becomes to the causal \textit{intervention}-based pretraining. 
We note that our goal is not performing theoretically causal inference but learning generic and de-biased visio-linguistic representations that can well generalize to downstream tasks with unknown data distributions.

We are particularly targeting at the Bert-style pretraining models and the context-based proxy tasks for supervision, such as masked language/object modeling (MLM/MOM). Context-based proxy tasks solely care about association, \ie, what co-occur with the anchor token without considering whether there are spurious correlations (\eg, \texttt{shirt} cannot cause \texttt{instrument}, and vice versa) or not. More formally, masked token modeling, abbreviated as MTM, models the conditional probability $P(Y|X)$ as the distribution of $Y$ when observing $X$. $Y$ is the masked token and $X$ denotes the context information. The spurious correlation occurs when $X$ and $Y$ are confounded by a common cause $Z$, as depicted in Figure \ref{fig:first}. Our goal is to model the interventional operation $P(Y|do(X))$, meaning the distribution of $Y$ when controlling $X$ to mitigate the correlation bias as we introduced before. 
Real-world cases concerning the conditional probabilities and the corresponding intervention results from the Conceptual Captions dataset can be found in Figure \ref{fig:condi}. In this paper, we propose several intervention-based BERT architectures to help learn deconfouned visio-linguistic representations. We name this kind of architectures as \textbf{DeVLBert}, which refers to \textit{\textbf{De}confounded \textbf{V}isio-\textbf{L}inguisitic \textbf{Bert}}. DeVLBert is designed as model-agnostic and can be easily encapsulated into any other Bert-style models.

We conduct in-depth experiments to discuss the performance of the proposed DeVLBert architectures. 
Pretraining is performed on the Conceptual Caption dataset which most downstream tasks are not built on, \ie, out-of-domain dataset. We evaluate the effects of these architectures on three downstream cross-modal tasks, including text-to-image retrieval \cite{Wang_Li_Lazebnik_2016}, zero-shot text-to-image retrieval, and visual question answering \cite{Antol_Agrawal_Lu_Mitchell_Batra_Zitnick_Parikh_2015}. We also conduct case studies to evaluate DeVLBert from the human perspective, and demonstrate that mitigating dataset biases boosts the generalization ability.

The main contributions of our work are summarized as follows:

\begin{itemize}
	\item We investigate the problem of out-of-domain pretraining, where pre-trained models are transferred to downstream tasks with unknown data distributions.
	\item We propose the novel DeVLBert framework designed for the Bert-style pretraining architectures with causal intervention to mitigate the spurious correlations caused by the context-based proxy tasks.
	\item We devise four implementations of the DeVLBert framework \footnote{Code will be released at \url{https://github.com/shengyuzhang/DeVLBert}} and discuss the empirical performance on several downstream tasks. The advantages of the DeVLBert are demonstrated by quantitative experiments, ablation studies, and case studies.
\end{itemize}

\section{Related Works}

\subsection{Visio-linguistic Pretraining}

Visio-linguistic (cross-modal) pretraining is a nascent research area that attracts considerable interests in recent years due to their strong ability of knowledge transfer. 
Existing works \cite{Li_Yatskar_Yin_Hsieh_Chang_2019,Lu_Batra_Parikh_Lee_2019,Su_Zhu_Cao_Li_Lu_Wei_Dai_2020,Chen_Li_Yu_Kholy_Ahmed_Gan_Cheng_Liu_2019,Li_Duan_Fang_Jiang_Zhou_2019,Tan_Bansal_2019} are mainly based on the Bert framework, including single-stream models and two-stream models. Besides the model structure, these methods differ mainly in the pretraining datasets and proxy tasks. 
Considerable methods incorporate in-domain datasets for pretraining, such as MSCOCO \cite{Lin_Maire_Belongie_Hays_Perona_Ramanan_Dollr_Zitnick_2014} and Visual Genome \cite{Krishna_Zhu_Groth_Johnson_Hata_Kravitz_Chen_Kalantidis_Li_Shamma_et_2017}. , which means the datasets are shared in the processes of both pretraining and downstream task training. 
In this paper, we investigate the particular out-of-domain pretraining problem, which can be a more general setting in the real-world.

\subsection{Causality in Vision \& Language}

It is of increasing research interests in computer vision that attempts to borrow useful analysis tools from causality. Typical works are concerning object tracking \cite{Lebeda_Hadfield_Bowden_2015,Xu_Qin_Liu_Xie_Zhu_2018}, interpretable Learning \cite{Kim_Canny_2017}, image classification \cite{Chalupka_Perona_Eberhardt_2015,Lopez-Paz_Nishihara_Chintala_Schlkopf_Bottou_2017}, and image generation \cite{Kocaoglu_Snyder_Dimakis_Vishwanath_2018,Bau_Zhu_Strobelt_Zhou_Tenenbaum_Freeman_Torralba_2019}. More recently, Wang \etal \cite{Wang_Huang_Zhang_Sun_2020} proposes the VC R-CNN framework for visual representation learning. They employ the causal intervention to deal with spurious correlation within datasets for visual common sense learning. However, VC R-CNN solely concerns intervention for visual domain. Alleviating the spurious correlations between vision and language in visio-linguistic pretraining can be also necessary, especially for cross-modal downstream tasks.

There are also considerable works that explore causality in natural language processing, varying among relation identification \cite{Ning_Feng_Wu_Roth_2018}, text classification \cite{Wood-Doughty_Shpitser_Dredze_2018}, and question answering \cite{Sharp_Surdeanu_Jansen_Clark_Hammond_2016}. 
More recently, some causality-related techniques also emerge in the research field of cross-modality. For example, visual dialogue \cite{Qi_Niu_Huang_Zhang_2019}, image caption \cite{Yang_Zhang_Cai_2020}, scene graph generation \cite{Tang_Niu_Huang_Shi_Zhang_2020} and VQA \cite{niu2020counterfactual}. Different from these works, DeVLBert investigates generic representation learning.

\section{Visio-Linguistic Bert}

\subsection{Bidirectional Transformer} \label{sec:BERT}

We choose Bidirectional Transformer (Bert) as our backbone structure, which can be the main-stream pretraining architecture for both natural language pretraining and visio-linguistic pretraining. For brevity, we take natural language pretraining as an example to illustrate the Bert structure. Given a language sequence $ S = \{ w_t \}_{t=1,\dots,N_w}$, Bert is pretrained to produce the contextualized word representations $ \mathbf{S} = \{ \mathbf{w}_t \}_{t=1,\dots,N_w} $. We note that the first token is often a pre-defined "[CLS]" and the learned representation for this token denotes the global representation of the whole sequence. Bert is composed of $N_l$ Transformer layers where each of the layers will output a feature sequence $ \mathbf{S}^l = \{ \mathbf{w}^l_t \}_{t=1,\dots,N_w} $.  We view the feature sequence of the last layer as the final pretrained representations, \ie, $\mathbf{w}_t = \mathbf{w}^{N_l}_t$. Each Transformer layer consists of a multi-head self attention module and a feed-forward module. Formally, such a process can be formulated as:
\begin{align}
    \mathbf{w}^l &= \textrm{MultiHeadAttention}(\mathbf{w}^{l-1}), \\
    \tilde{\mathbf{w}}^l &= \textrm{LayerNorm}(\mathbf{w}^{l-1} + \mathbf{w}^l), \\
    \hat{\mathbf{w}}^l &= \textrm{FeedForward}(\tilde{\mathbf{w}}^l), \\
    \mathbf{w}^l &= \textrm{LayerNorm}(\tilde{\mathbf{w}}^l + \hat{\mathbf{w}}^l).
\end{align}
Each self attention head connect all pairs of input and output positions \cite{Vaswani_Shazeer_Parmar_Uszkoreit_Jones_Gomez_Kaiser_Polosukhin_2017}:
\begin{align}
\textrm{ SelfAttention }(\mathbf{S})=\operatorname{softmax}\left(\frac{Q(\mathbf{S}) K(\mathbf{S})^{T}}{\sqrt{D_{w}}}\right) V(\mathbf{S}),
\end{align}
while $Q$ and $K$ are learnable query transformation and key transformation to compute the attention weights. $V$ is the learnable linear transformation to obtain the value context. $D_{w}$ is the feature dimension of keys, values, and queries. This design yields a global correlation, \ie, the final representation of each word token will be correlated with all other words. Since this modeling schema has no sense of word order (in the sequence) unlike RNNs, it is necessary add the position signals onto each word embedding. There are some techniques doing this, such as position embedding \cite{Gehring_Auli_Grangier_Yarats_Dauphin_2017} and position encoding \cite{Vaswani_Shazeer_Parmar_Uszkoreit_Jones_Gomez_Kaiser_Polosukhin_2017}, and we follow the position embedding for simplicity. For the $t$th word token, the initial representation can be $\mathbf{w}_t^0 = \mathbf{w}_t^e + p_t$, where $\mathbf{w}_t^e$ denotes the word representation taken from the embedding layer and $p_t$ denotes the embedding for the position index $t$.

This structure can be easily transferred to the visual domain if the visual words are reasonably defined. A simple yet effective approach is to view the sub-regions of interest as visual words. More concretely, object detectors, such as Faster-RCNN \cite{Ren_He_Girshick_Sun_2015}, will be used to extract object bounding boxes and object feature maps. The original object representations $\mathbf{O}^e = \{ \mathbf{o}^e_i \}_{i=1,\dots,N_v}$ are obtained by global average pooling over the object feature maps. Similar to the language pretraining, we add a global representation $\mathbf{o}^e_{[\textrm{G}]} = 1/N_v \sum_{i=1}^{N_v}\mathbf{o}^e_i$ at the beginning of the sequence. To be aware of the position signals of objects, we encode the information within bounding boxes to obtain position encodings. Concretely, each bounding box can be represented as a 5-d vector, including the normalized top-left coordinates, the normalized bottom-right coordinates, and the scaling factor. We note that for the global representation, the bounding box refers to that of the entire image. The position encoding vector of the same dimension as the object representation is then obtained by a feed-forward network.

\subsection{Two-stream Visio-Linguistic Modeling}

Existing Bert-like visio-linguistic pretraining methods can be roughly categorized into single-stream architectures and two-stream architectures. We are following the two-stream architectures \cite{Lu_Batra_Parikh_Lee_2019,Su_Zhu_Cao_Li_Lu_Wei_Dai_2020,Tan_Bansal_2019} since they keep the independence of each modality as well as modeling the interaction across different modalities. The only difference between the visio-linguistic Transformer layer and the modality-specific Transformer layer lies in the queries. Formally, the attention head in visio-linguistic Transformer layer for the language side can be formulated as:
\begin{align}
\textrm{ SelfAttention }(\mathbf{S}, \mathbf{O})=\operatorname{softmax}\left(\frac{Q(\mathbf{S}) K(\mathbf{O})^{T}}{\sqrt{D_{w}}}\right) V(\mathbf{O}),
\end{align}
Intuitively, this process is designed to borrow language-related information (language features $\textbf{S}$ as queries) from the visual features ($\textbf{O}$ as keys and values). Likewise, for the visual side, we have:
\begin{align}
\textrm{ SelfAttention }(\mathbf{O}, \mathbf{S})=\operatorname{softmax}\left(\frac{Q(\mathbf{O}) K(\mathbf{S})^{T}}{\sqrt{D_{o}}}\right) V(\mathbf{S}).
\end{align}
where $D_{o}$ denotes the feature dimension of visual queries. By combining the visio-linguistic Transformer layer and the modality-specific Transformer layer and further stacking the combined layers, we obtain the main structure of two-stream visio-linguistic Bert.

%--------------------------------fig---------------------
\begin{figure*}[t] \begin{center}
    \includegraphics[width=0.8\textwidth]{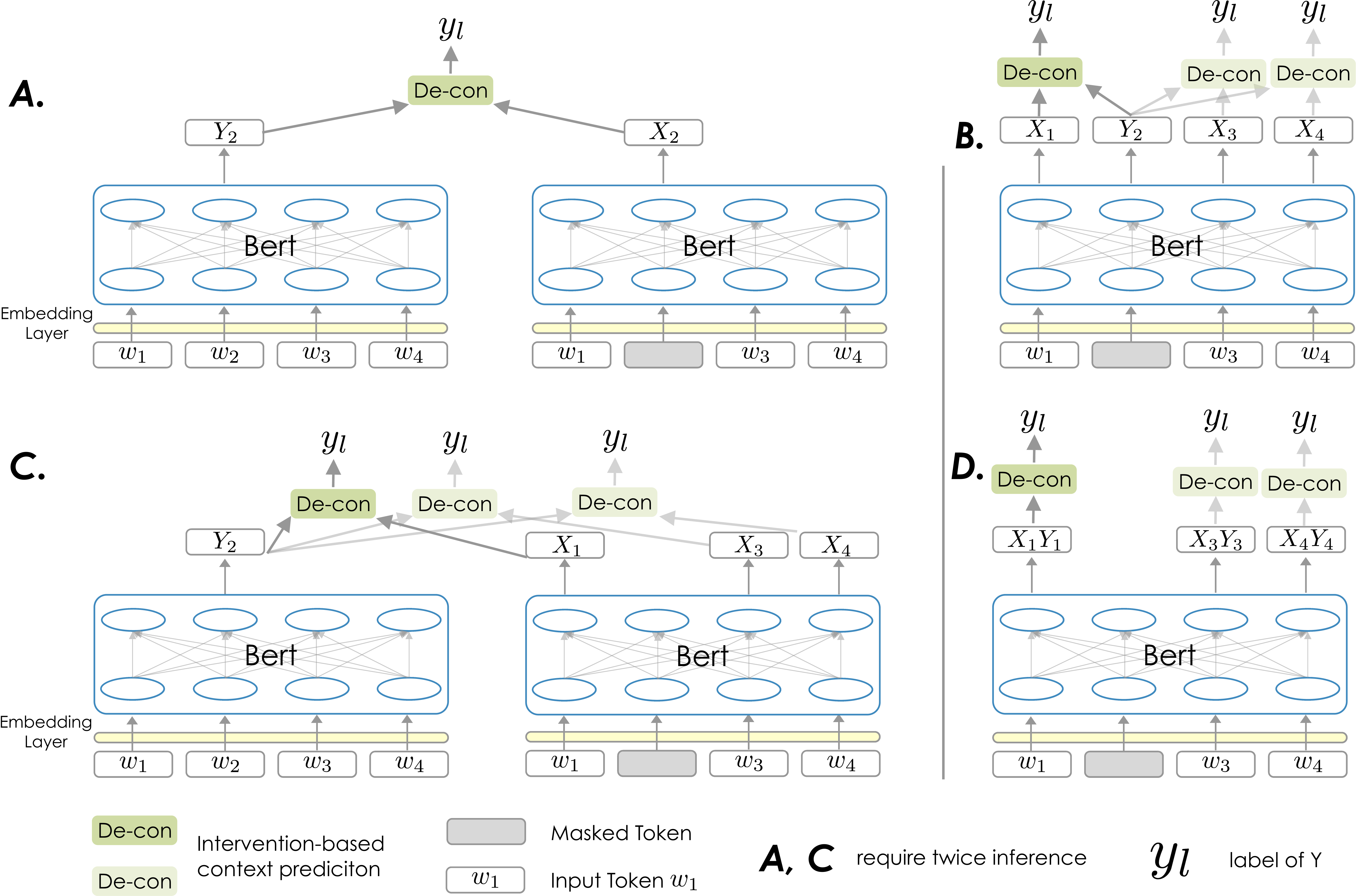}
    \caption{
    A vivid illustration of four intervention formulations for Bert-style training. Deciding the forms of X and Y in Bert is essential for further intervention-based context prediction. Design A\&C require twice inference. Design A\&B replace the frequently used masked-token-modeling (MTM) objective while Design C\&D are independent of the MTM.
	}
\vspace{-0.2cm}
\label{fig:imple}
\end{center} \end{figure*}
%--------------------------------fig end---------------------

\subsection{Pretraining Proxy Tasks}

Masked language modeling (MLM) and masked object modeling (MOM) are popular context-based proxy tasks for language and vision, respectively. We group MLM and MOM into masked token modeling (MTM) in this paper for brevity. As the name (masked) implies, MTM often randomly masks tokens in a sequence by a given probability and replace them with special tokens, such as "[MASK]". Then, Bert is required to make predictions to recover these tokens. Typical recovering strategies include token vocabulary classification and feature regression. For MLM, we follow the standard practice in original BERT\cite{Devlin_Chang_Lee_Toutanova_2019}. For MOM, 15\% proportion of the objects will be masked ready for recovering. Since no objects are masked during testing, which yields a setting gap between training and testing, we replace 10\% of the masked objects with their original representations. We note that these objects are still required to be \textit{recovered}, which are different from the initially unmasked objects. By recovering, we mean the object classification with soft labels, which come from the object detectors.

Visio-linguistic alignment follows the original design of the next sentence prediction objective in natural language pretraining. By the probability of 50\%, the originally paired sentence will be replaced by a random sampled unpaired sentence. The Bert model is required to predict whether the language sequence and the objects sequence are aligned. We employ a simple feed-forward neural network to compute the alignment score based on the global representation of words/objects sequence, \ie, $\mathbf{w}_{[CLS]}$ and $\mathbf{o}_{[G]}$, where $\mathbf{o}_{[G]}$ denotes the final representation of the special global object $o_{[G]}$.

\section{Deconfounded Vsio-Linguistic Bert}

\subsection{Bert in the causal view}

As illustrated in section \ref{sec:BERT}, the Transformer layer connects each output token representation with all input token representations. We denote the representation of one output token as $Y$ and the representations of all other tokens as $X$. Bert models the function of $P(Y|X)$. In the causal view, there can be some confounder $Z$ affecting both $X$ and $Y$. If such confounders are not controlled in modeling, some false conclusion about $X$ and $Y$ may be drawn as part or all of the effect might come from $Z$.
The key to alleviate the spurious correlations is to control the confounders when evaluating the causal effect of $Y$ given $X$, \ie, intervention-based modeling $P(Y|do(X))$. Our framework borrows the idea of backdoor adjustment \cite{neuberg2003causality, Wang_Huang_Zhang_Sun_2020}. Formally, by the Bayes Rule, the conventional likelihood can be re-written as:
\begin{align}
	P(Y | X) = \sum_{z} P(Y, z | X) = \sum_{z} P(Y | X, z) \underline{P(z | X)},
\end{align}
By using the \textit{do}-calculus, we remove any incoming influence to the intervened variable, \ie, $X$. By the definition of \textit{do}-calculus, we have:
\begin{align}
	P(Y | do(X)) &= \sum_{z} P(Y, z | do(X)) \\
	 &= \sum_{z} P(Y | do(X), z) P(z | do(X)) \\
	 &= \sum_{z} P(Y | X, z) \underline{P(z)}.
\end{align}
The proof of transitions $ P(Y | do(X)) = P(Y | X, z)$ and $P(z | do(X) = P(z)$ can be found in the book \cite{Aronow_Svje_2020} and in several following works \cite{Wang_Huang_Zhang_Sun_2020, Qi_Niu_Huang_Zhang_2019}. The prior probability of each $z$ can be easily pre-counted before training following \cite{Wang_Huang_Zhang_Sun_2020}. It is infeasible to individually model the distribution of $P(Y|X,z)$ for each $z$ as the number of potential confounders can be large. We borrow the idea of Normalized Weighted Geometric Mean \cite{ Xu_Ba_Kiros_Cho_Courville_Salakhutdinov_Zemel_Bengio_2015,Srivastava_Hinton_Krizhevsky_Sutskever_Salakhutdinov_2014} to approximate the expensive sampling and separate modelling as \cite{Wang_Huang_Zhang_Sun_2020} does. Formally, if the last objective is classification, we can re-write the following terms:
\begin{align}
	P(Y | X, z) &= \textrm{softmax}\left(f_c(\mathbf{x}, \mathbf{z})\right), \\
	P(Y | do(X)) &= \mathbb{E}_{z}\left[\operatorname{softmax}\left(f_{c}(\mathbf{x}, \mathbf{z})\right)\right],
\end{align}
where $\mathbf{x}$ and $\mathbf{z}$ denote the feature representations of $X$ and $z$, and $f_c$ denotes the classification head of intervention. The essence of NWGM is to move the expectation into the operation of softmax:
\begin{align}
	\mathbb{E}_{z}\left[\operatorname{softmax}\left(f_{c}(\mathbf{x}, \mathbf{z})\right)\right] \stackrel{NWGM}{\approx} \operatorname{softmax}\left(\mathbb{E}_{z}\left[f_{c}(\mathbf{x}, \mathbf{z})\right]\right).
\end{align}
%(\mathbf{W}_yy)^T(\mathbf{W}_zz)
In this paper, we model the term $f_{c}(\mathbf{x}, \mathbf{z})$ by the feed-forward neural network $\mathbf{W}_c[\mathbf{x}, \alpha_y(\mathbf{z}) * \mathbf{z}]$, where $[,]$ denotes the concatenation operation and $\alpha_y(\mathbf{z})$ denotes the importance factor that is parameterized by $\mathbf{y}$. The introduction of y-dependent confounder importance re-weighting strategy follows VC R-CNN \cite{Wang_Huang_Zhang_Sun_2020}. This formulation is reasonable in the sense that a particular $z$ that well correlates with $Y$ have high probability to be the confounder of $Y$ and $X$. Formally, we have:
\begin{align}
	\alpha_y(\mathbf{z}) &= \frac{(\mathbf{W}_y\mathbf{y})^T(\mathbf{W}_z\mathbf{z})}{\sum_{v \ne \varsigma} (\mathbf{W}_y\mathbf{y})^T(\mathbf{W}_z\mathbf{v})}, \\
	P(Y|do(X)) &= \operatorname{softmax} \left(\mathbf{W}_c\left[\mathbf{x}, \sum_z P(\mathbf{z}) * \alpha_y(\mathbf{z}) * \mathbf{z} \right]\right).
\end{align}
where $\mathbf{y}/\mathbf{v}$ is the feature representation of $Y/v$. $\varsigma$ denotes the confounder that has the same token class as $Y$. For example, if the predicted token is $y=\texttt{cat}$, it is unreasonable to take $z=\texttt{cat}$ as a potential confounder for predicting $y=\texttt{cat}$. We note that the corresponding weight $\alpha_y(\mathbf{\varsigma})$ is thus $0$. Now the problem of how to perform intervention-based learning is transformed to define how Bert models the feature representation of $X$ and $Y$. In this paper, we propose several implementations. We note that in the following illustrations, each case is concerning only one masked token, and it is easy to extend the framework to the general case with multiple masked tokens.

\begin{itemize}
	\item \textbf{Design A}. We firstly investigate how to harness masked token modeling with intervention, since 1) MTM is among the most popular pretraining proxy tasks. 2) MTM is solely based on likelihood estimation, which might introduce spurious correlations. Still, we take natural language pretraining as an example for illustration. For one masked word $w_t$, it is intuitively to view the final representation $\mathbf{w}_t$ as $\mathbf{x}_t$ since $\mathbf{w}_t$ contains no explicit information from the word itself (being masked). However, it is not easy to find $\mathbf{y}$ which contains the information of word $w_t$ in single inference. We choose to run another inference with no masked tokens. In this way, the final representation of word $w_t$ can be viewed as $\mathbf{y}$. This implementation is depicted in Figure \ref{fig:imple} A.
	\item \textbf{Design B}. Figure \ref{fig:imple} B depicts another design to harness MTM. Under the framework of MTM, Bert leverages and aggregates $w_t$-related information from the context to predict the label of $w_t$. In this perspective, the final representation of the masked token $w_t$ can be viewed as $\mathbf{y}_t$ while the final representations of all unmasked tokens can be viewed as $\{ \mathbf{x}_k \}_{k=1,\dots,t-1,t+1,\dots,N_w}$. This design is efficient without an extra inference process. The time complexity is $O(N_u * N_m)$, where $N_u$ and $N_m$ are the numbers of unmasked tokens and masked tokens, respectively.
	\item \textbf{Design C}. As depicted in Figure \ref{fig:imple} C, Design C is a variant of Design A and views the final representations of all unmasked tokens as $\{ \mathbf{x}_k \}_{k=1,\dots,t-1,t+1,}$${}_{\dots,N_w}$.
	\item \textbf{Design D}. By viewing the final representations of unmasked tokens as integrated representations of $X$ and $Y$, Design D is non-intrusive and can be the most efficient among the proposed designs. By non-intrusive, we mean that this design makes fewer modifications, \ie, without another forward run and without hurting the original MTM objective. With the time complexity of $O(N_u)$, Design D is more efficient than the double-run designs as well as Design B. We note that in this design, the modeling of $P(Y|do(X))$ is slightly different:
\begin{align}
	\alpha_r(\mathbf{z}) &= \frac{(\mathbf{W}_r\mathbf{r})^T(\mathbf{W}_z\mathbf{z})}{\sum_z (\mathbf{W}_r\mathbf{r})^T(\mathbf{W}_z\mathbf{z})}, \\
	P(Y|do(X)) &= \operatorname{softmax} \left(\mathbf{W}_c\sum_z P(\mathbf{z}) * \alpha_r(\mathbf{z}) * \mathbf{z} \right).
\end{align}
where $\mathbf{r}$ denotes the integrated representation of $\mathbf{y}$ and $\mathbf{x}$, and $\alpha_r(\mathbf{z})$ is the importance factor parameterized by $\mathbf{r}$. Since the representation $\mathbf{x}$ is no longer available, we omit the concatenation operation. Using both $\mathbf{y}$ and $\mathbf{x}$ to re-weight the importance of each $\mathbf{z}$ is also reasonable in the sense that a particular $z$ that well correlates with both $Y$ and $X$ have high probability to be the confounder of $Y$ and $X$.

\end{itemize}

%Since masked token modeling (MTM) is among the most popular pretraining proxy tasks, we firstly investigate how to harness MTM with 
%

\subsection{Intra- \& Inter-modality Intervention}

\vpara{Vision deconfounding \& Vision Confounder Set.} It is infeasible to take each particular object (in some image) as a potential confounder as there can be numerous objects in pretraining datasets. Following VC R-CNN \cite{Wang_Huang_Zhang_Sun_2020}, we consider the high-level object classes as potential confounders. The representation of each object class is obtained by averaging pooling the set of object features belonging to the class (but in different images). The size of the confounder set (1,600) is equivalent to the number of object classes that are pre-defined by the pre-trained object detector. For vision deconfounding, $Y$ and $X$ are only selected from the final representations of the visual regions, and confounders in the vision confounder set are discussed. For Design A and B, the MOM objective is totally replaced by the intervention objective. For Design C and D, the intervention is married with the MOM objective.

\vpara{Language deconfounding \& Language Confounder Set.} Likewise, it is infeasible to take each particular word token (in some sentence) as a potential confounder, and it is also expensive to discuss all high-level words since there are about 30,000 words in the Bert vocabulary. In this paper, we choose nouns as potential confounders since 1) nouns are content words that have meaning or semantic value \cite{Yang_Liu_2020}; 2) the role of nouns is similar to the role of objects in image, which might ease the inter-modality intervention. Specifically, we use the NLTK toolkit \cite{Bird_2006} to perform Part-of-Speech Tagging, and choose word tokens of which the tags belong to $["NN", "NNS", "NNP", "NNPS"]$ as potential confounders. There are in total 156 potential confounders in the language confounder set (after removing nouns with low-frequencies since such words have less chance to be confounders). The feature representation of each noun is initialized as the mean-pooled vector of the Bert contextual embeddings of words (the same noun) in different sentences. Similarly, for language deconfounding, $Y$ and $X$ are only selected from the final representations of the language words, and confounders in the language confounder set are discussed. We note that only noun words are considered as $X$ and $Y$ since they are of high probability to pose spurious correlations with visual objects. For Design A and B, the MLM objective is replaced by the intervention objective for masked noun words, and others words still have chance to process masked prediction. For Design C and D, the masked prediction objective is not affected by the intervention objective.

\vpara{Inter-modality Intervention.} It is necessary to perform inter-modality (or cross-modal) intervention since with the two-stream visio-linguistic modeling, the token representations of each modality contain information from other modalities, which may lead to spurious correlations without inter-modality intervention. We observe that in the Conceptual Caption dataset, the conditional probability of visual object "\texttt{shirt}" give the word "\texttt{instrument}" is about 6\% while \texttt{shirt} and \texttt{instrument} have no causal relationship but might have a common cause, \ie, the visual object or the language word \texttt{person}. Specifically, for inter-modality intervention, $Y$ and $X$ can be tokens from different modalities, and confounders can be selected from both vision and language confounder sets.

We note that both the MTM and the intervention-based objective conduct object classification for vision, and word prediction for language. The difference between intervention and MTM is that intervention discusses the effect of $Y$ given $X$ and each potential confounder $z$, which helps mitigate the spurious correlations.

%--------------------------------table---------------------
\begin{table*}[t]
%\begin{strip}
\centering
\caption{
     Comparison between DeVLBert and other competitors, including ViLBERT which only uses out-of-domain${}^\circ$ pretraining datasets, VisualBERT only uses in-domain${}^\bullet$ datasets, and InterBert using both${}^\odot$.
}
\setlength{\tabcolsep}{6.5pt}
\setlength\doublerulesep{0.5pt}
\begin{tabular}{l|ccc|ccc|cc}
& \multicolumn{3}{c}{Image Retrieval (IR)} & \multicolumn{3}{c}{Zero-shot IR} & \multicolumn{2}{c}{VQA} \\ 
\multicolumn{1}{c|}{Methods}    & R@1    & R@5   & R@10    & R@1    & R@5   & R@10     & test-dev   & test-std    \\  
\hline \hline
SCAN  \cite{Lee_Chen_Hua_Hu_He_2018}        & 48.6 & 77.7 & 85.2 & - & - & - & - & -   \\ %   & - & - &  -       \\
BUTD  \cite{Anderson_He_Buehler_Teney_Johnson_Gould_Zhang_2018}  &  - & - & - & - & - & - & 65.3 & 65.7    \\
\hline
${}^\bullet$VisualBERT \cite{Li_Yatskar_Yin_Hsieh_Chang_2019}     & - & - & - & - & - & - & 70.8 & 71.0  \\ 
${}^\odot$InterBert \cite{Lin_Yang_Zhang_Liu_Zhou_Yang_2020}   & 61.9 & 87.1 & 92.7 & 49.2 & 77.6 & 86.0 & 70.3 & 70.6   \\ 
\hline
${}^\circ$ViLBERT \cite{Lu_Batra_Parikh_Lee_2019} (Baseline)    & 58.2 & 84.9 & 91.5 & 31.9 & 61.1 & 72.8 & 70.6 & 70.9   \\ 
${}^\circ$DeVLBert   & 61.6 & 87.1 & 92.6 & 36.0 & 67.1 & 78.3 & 71.1 & 71.5 \\ 
\hline

\end{tabular}
\label{tab:quantitative}
\vspace{-0.2cm}
\end{table*}
%
%--------------------------------table end---------------------

\section{Experiments}

\subsection{Experiment Setup}

\vpara{Pretraining DeVLBert.} We follow ViLBERT \cite{Lu_Batra_Parikh_Lee_2019} to pretrain DeVLBert on the Conceptual Caption \cite{Sharma_Ding_Goodman_Soricut_2018} dataset, which is an out-of-domain dataset that has little data overlap with most downstream tasks. Images and raw descriptions are harvested from HTML pages that contain images and Alt-text attributes. Then, automatic language cleaning pipelines are developed to obtain the final image captions that are clean, informative, but less similar to the human-annotated captions in datasets of downstream tasks. In other words, the Conceptual Caption dataset serves as an excellent dataset for out-of-domain pretraining. Due to broken or expired links by the time we downloaded, we use around 3.04 million <image, caption> pairs for pretraining, which is smaller than the original 3.3 million dataset when first published, and also smaller than the 3.1 million dataset used in ViLBERT. To make our results comparable to the previous out-of-domain pretraining work, \ie, ViLBERT, we are following the exact pipeline as theirs, including the initialization of the linguistic stream and visual region feature extraction.

\vpara{Finetuning on downstream tasks.} Also, we are following the pipelines of three downstream tasks, \ie, Text-to-Image Retrieval (IR), Zero-shot Text-to-Image Retrieval (Zero-shot IR), and Visual Question Answering (VQA) of ViLBERT. For more details, such as dataset split, fine-tuning strategies, and hyper-parameters, please refer to ViLBERT\cite{Lu_Batra_Parikh_Lee_2019}. We note that our goal is not achieving the state-of-the-art with bells\&whistles but demonstrating the effectiveness of mitigating spurious correlations of DeVLBert for out-of-domain pretraining. Besides the quantitative evaluations, we conduct user studies that qualitatively show whether and how DeVLBert achieves better results by mitigating biases.

%--------------------------------table---------------------
\begin{table}[!t]
%\begin{strip}
\centering
\caption{
    Comparisons between different DeVLBert implementations, and ablation studies on the architecture D.
}
\setlength\doublerulesep{0.5pt}
%\footnotesize  
%\vspace{-0.1cm}
%\small
\begin{tabular}{l|ccc|ccc}
%\hline
%\toprule
& \multicolumn{3}{c}{Image Retrieval (IR)  } & \multicolumn{3}{c}{Zero-shot IR}  \\
%\cline{1-9}
% <<<
\multicolumn{1}{c|}{Method}          & R@1    & R@5   & R@10    & R@1    & R@5   & R@10        \\
%\midrule
\hline \hline
Baseline & 58.2 & 84.9 & 91.5 & 31.9 & 61.1 & 72.8
\\
\hline
A-V         & 60.3	 & 86.24 & 92.06 & 30.18 & 59.46 & 71.88        
\\
A-VL         & 58.3 & 85.5 & 91.6 & 25.4 & 54.7 & 67.2        
\\
\hline
B-V     & 58.9  & 85.3  & 91.1   & 33.0 & 62.2 & 74.0 \\
\hline
C-V         & - & - & - & 27.0 & 56.2 & 69        
\\   
\hline    
D-V         &  59.3 & 85.4 & 91.8 & 32.8 & 63.0 & 74.1      
\\     
D-VL         & 60.3 & 86.7 & 92.2 & 34.9 & 65.5 & 77.0
      
\\    
D-VLC         & \textbf{61.6} & \textbf{87.1} & \textbf{92.6} & \textbf{36.0} & \textbf{67.1} & \textbf{78.3}       
\\
\hline
% >>>
\end{tabular}
%\medskip
\vspace{-0.2cm}
\label{tab:designs}
\end{table}

%--------------------------------table end---------------------

\vpara{Hardware \& Software Configuration} We implement the models in python3.6 and PyTorch 1.1.0 \cite{paszke2017automatic}, and train the models on a Linux server equipped with 8 NVIDIA V100-SXM2-16GB GPUs.

\subsection{Quantitative Evaluation}

By quantitative evaluation, we care about a few issues listed below:

\vpara{How do different intervention-based architectures perform?} To answer this question, we evaluate the performance of different architectures on the downstream tasks, \ie, image retrieval, and zero-shot image retrieval. The results are listed in Table \ref{tab:designs}. We use A-V to denote the architecture of design A,  A-VL to denote the architecture of design A with both vision and language deconfounding. Based on the results, we can see that: 

\begin{itemize}
	\item Most of the architectures obtain performance gain on at least one of the tasks, which demonstrates the effectiveness of intervention-based learning.
	\item The twice inference design achieves inferior results on the zero-shot image retrieval task. Partially due to the complexities introduced by another inference, it might take more iterations to converge, which can be expensive. Moreover, similar to the results shown in VC R-CNN \cite{Wang_Huang_Zhang_Sun_2020} that the performance of directly using the pre-trained commonsense features (intervention-based) is lower than that of the original features (association-based) while combining these two features would achieve the best performance. In our case, combining the pretrained deconfounding features with the knowledge in the downstream task (regular image retrieval) achieves better results in zero-shot IR.
	\item Comparing A-VL with A-V, the introduction of language deconfounding leads to a performance drop on IR and zero-shot IR. We attribute this phenomenon to the incomplete training of MTM. For the language side, following ViLBERT, the classification module shares the word embedding matrix with the input embedding layer. For A-VL, we only mask noun words since the language confounder set comprises only noun words. Therefore, the embedding matrix solely sees noun words in the classification, which leads to inferior results due to incomplete learning of other words. Non-intrusive design D mitigates this problem.
	\item Without the structure and training complexities introduced by the other inference, B-V and D-V show clear advantages over A-C and C-V.
	\item D-V further outperforms the architecture of B-V, and we attribute this consistent improvement to the non-intrusive intervention modeling. More concretely, isolating the masked token modeling makes the shared embedding module in the MTM classification module learn better. Meanwhile, architecture D is the most efficient.
\end{itemize}

%--------------------------------table---------------------
\begin{table}[!t]
%\begin{strip}
\centering
\caption{
    Comparisons between DeVLBert and VC R-CNN (with Bert as language feature extractor).
}
\setlength\doublerulesep{0.5pt}
\begin{tabular}{l|ccc|cc}
& \multicolumn{3}{c}{Image Retrieval (IR)  } & \multicolumn{2}{c}{VQA}  \\
\multicolumn{1}{c|}{Method}          & R@1    & R@5   & R@10    & test-dev    & test-std        \\
\hline \hline
VC R-CNN \cite{Wang_Huang_Zhang_Sun_2020}         & 15.0 & 40.4 & 54.7 & 53.2 & 53.6     
\\
\hline
DeVLBert         & 39.2 & 70.2 & 80.5 & 53.5 & 53.9
\\
\hline
\end{tabular}
\vspace{-0.2cm}
\label{tab:vcrcnn}
\end{table}

%--------------------------------table end---------------------

%--------------------------------fig---------------------
\begin{figure*}[t] \begin{center}
    \includegraphics[width=0.9\textwidth]{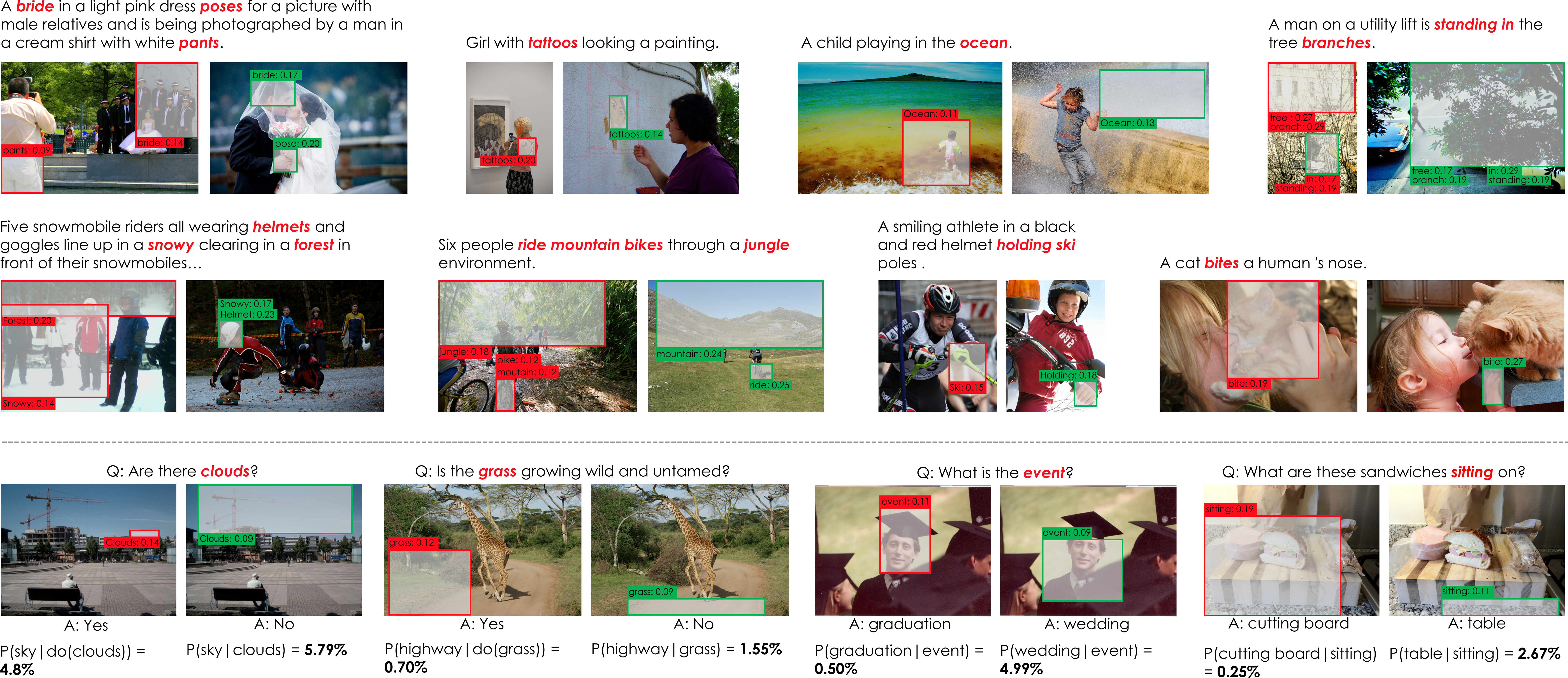}
    \caption{
    Case studies by visualizing the attention of the last cross-modal attention layer in DeVLBert (the left for each case) and ViLBERT (right). Cases are sampled from the testing/validation set of downstream tasks, \ie, image retrieval (top), and VQA (bottom). The labeled word is the attention query word from the input sentence, and the number is the corresponding attention weight.
	}
\vspace{-0.2cm}
\label{fig:cases}
\end{center} \end{figure*}
%--------------------------------fig end---------------------

\vpara{Do both intra-modality intervention and inter-modality intervention improve the out-of-domain pretraining?} Since architecture D-V achieves the best performance, we further extend architecture D-V to architecture D-VL by incorporating language deconfounding, and architecture D-VLC by incorporating the cross-modal (inter-modality) deconfounding. The evaluation results are shown in Table \ref{tab:designs}, it can be seen that removing any deconfounding component will lead to a performance drop, which again verifies the effectiveness of the proposed framework.

\vpara{How does DeVLBert perform comparing with existing visio-linguistic pretraining methods and task-specific downstream SOTA models?} We view the architecture D-VLC as DeVLBert due to its empirical effectiveness. As shown in Table \ref{tab:quantitative}, when compared to task-specific SOTA models, including SCAN \cite{Lee_Chen_Hua_Hu_He_2018} for image retrieval, BUTD \cite{Anderson_He_Buehler_Teney_Johnson_Gould_Zhang_2018} for visual question answering, DeVLBert yields a large margin improvement over these methods. More importantly, while the baseline method, \ie, ViLBERT \cite{Lu_Batra_Parikh_Lee_2019}, cannot beat in-domain pretraining methods, including VisualBERT \cite{Li_Yatskar_Yin_Hsieh_Chang_2019} using MSCOCO as the pretraining dataset, and InterBert using both Conceptual Caption and MSCOCO as pretraining datasets, DeVLBert obtains a performance boost over the VisualBert and InterBert on the VQA task and achieves comparable results to the InterBert on the image retrieval task. On zero-shot image retrieval, DeVLBert cannot beat InterBert, which is a reasonable result since MSCOCO is the testing dataset for zero-shot image retrieval, and InterBert uses MSCOCO for pretraining. 

\begin{sloppypar}
	\vpara{How does DeVLBert perform comparing with VC R-CNN?} The comparison results with VC R-CNN \cite{Wang_Huang_Zhang_Sun_2020}, which learns visual commonsense features by intervention, are listed in Table \ref{tab:vcrcnn}. We use VC R-CNN as the visual feature extractor and vanilla Bert as the language feature extractor. For fair comparison, we do not fine-tune DeVLBert on the downstream datasets either, and view DeVLBert as visio-language feature extractor. We mean-pool the extracted features for each modality and concatenate the pooled features from two modalities. We use two layer MLP as classifier with hidden size twice as large as the input feature size. According to Table \ref{tab:vcrcnn}, DeVLBert achieves large performance gain over VC R-CNN on multi-modal matching tasks (Image Retrieval) and competitive results on VQA, which shows cross-modal pretraining and deconfounding are essential for cross-modal downstream tasks.
\end{sloppypar}

\subsection{Case Studies}

To further evaluate the effectiveness of DeVLBert from the human perspective, we follow \cite{Wang_Huang_Zhang_Sun_2020} to conduct case studies on the testing/validation set of downstream tasks, including image retrieval and VQA (See Figure \ref{fig:cases}). For image retrieval, given a query sentence, we select the top answer image of DeVLBert (left) and ViLBERT (right). 
Compared to the VC R-CNN that focusing on visual attention, here we are especially interested in cross-modal attention, which is essential for visual-language tasks. Concretely, there are multiple co-attention blocks in both ViLBERT and DeVLBert, and we select the last block to obtain task-oriented association (the closer to the classification layer, the more task-oriented). Each co-attention block still contains multiple attention heads, we take the average attention map of all heads for visualization. We select the box with the biggest attention weight for each word. The results indicate that: 1) \textbf{The attended visual tokens (object boxes) of DeVLBert are more accurate than those of ViLBERT.} By "accurate", we mean the attended tokens are more useful for determining whether this image is locally relevant to the query sentence, and better as reasoning cues given the question. For example, in the case $C_{34}$, which denotes the case in the $3$rd row and the $4$th column, the attended box of ViLBERT (right) directly focuses on \texttt{sitting} and fails to consider the \texttt{sandwiches sitting}, \ie, question-specific context, while the attended box of DeVLBert is more accurate. We further compute the conditional probability of the answer given word \texttt{sitting}, which shows that DeVLBert can generate less frequent but more accurate answers. 2) \textbf{The results of DeVLBert yields less cognitive errors or spurious correlations}. For example, in case $C_{11}$, ViLBERT considers "person with wedding veil" as the "bride", and view the man as "bride" by mistake. In case $C_{32}$, there is a spurious correlation between the word \texttt{grass} and the visual object \texttt{highway}, which drives the ViLBERT to attend to the region with both \texttt{grass} and \texttt{highway}. With such attended region, ViLBERT fail to realize that the grass is growing wild and untamed. The conditional probabilities under $C_{31}$ and $C_{32}$ show DeVLBert can learn to pay less attention to spuriously correlated tokens such as \texttt{sky} and \texttt{highway} by deconfounding.

\section{Conclusion}

In this paper, we propose to mitigate the spurious correlations for out-of-domain visio-linguistic pretraining. The fact that each output token is connected with all input tokens in Bert, and the pure association nature of masked token modeling objective makes the problem more severe. We borrow the idea of back-door adjustment to propose four novel Bert-style architectures as DeVLBert for out-of-domain pretraining. We conduct extensive quantitative evaluations as well as ablation studies to discuss the empirical effectiveness of different architectures. The results show that DeVLBert can achieve promising numerical results compared to the baseline and even some in-domain visio-linguistic pretraining methods. 

\section{ACKNOWLEDGMENTS}

\begin{sloppypar}
	The work is supported by the NSFC (61625107, 61751209, 61836002), National Key R\&D Program of China (No. 2018AAA0101900, No. 2018AAA0100603), Zhejiang Natural Science Foundation (LR19F020006), Fundamental Research Funds for the Central Universities (2020QNA5024),  and a research fund supported by Alibaba.
\end{sloppypar}

%In this paper, we

\bibliographystyle{ACM-Reference-Format}
\bibliography{9.citations}

\end{document}